**Letter to the Editor: What are the legal and ethical considerations of submitting radiology reports to ChatGPT?**

Siddharth Agarwal[1], David Wood[1], Robin Carpenter[1], Yiran Wei[1], Marc Modat[1], Thomas C Booth[1,2]

1. School of Biomedical Engineering & Imaging Sciences, King's College London, 9th Floor, Becket House, 1 Lambeth Palace Rd, London, SE1 7EU, UK
2. Department of Neuroradiology, Ruskin Wing, King's College Hospital NHS Foundation Trust, London SE5 9RS, UK

Correspondence to: Thomas C Booth, thomas.booth@kcl.ac.uk, +44 (0) 203 299 4828

To the Editor,

We read with great interest the article titled 'Large language models (LLMs) in the evaluation of emergency radiology reports: performance of ChatGPT-4, Perplexity, and Bard'.[1] The authors compare three LLMs' abilities to extract the presence of an urgent finding in emergency radiology reports. The study adds to a growing body of literature regarding how LLMs can perform classification tasks rather than natural language generation, the latter task being the one that they are best suited for. We would question the utility of LLMs to 'establish a reading priority based on the likely presence of emergency findings', given that a human radiologist is first required to generate the report from which the finding is extracted, however the task itself can be instrumental in generating large quantities of labels for downstream computer vision research tasks.[2–9]

We read with some concern that the authors felt that ethical approval was not required because 'patient-related information was not collected [and there was a] total absence of patient-identifiable data.' One should analogously consider if the use of patient imaging for research would require no ethical approval, even when the DICOM headers have been scrubbed and faces blocked out. In England, all research studies involving NHS patient data require Health Research Authority (HRA) approval, even if the data was fully anonymised. A Research Ethics Committee (REC) may share our concerns regarding the submission of patient reports to OpenAI and Google. Once submitted, these reports will be stored on proprietary servers located in the United States. Should a patient wish for data about them to be deleted, neither company would be able to identify that specific patient's data, nor would they be under any obligation to do so.

In addition, true anonymisation of patient reports, as opposed to pseudonymisation, may be difficult to achieve for some patients. Reidentification of anonymised patient data can sometimes be achieved by cross referencing the data with other datasets the patient also appears in.[10] Controlling the context the data sits in is therefore often important for maintaining privacy. The right to privacy is part of the 1950 European Convention on Human Rights and is a foundation for the General Data Protection Regulation (GDPR).[11] GDPR would apply if reports were pseudonymised rather than truly anonymised.

Data submitted to the non-enterprise version of ChatGPT is then used to train subsequent models and is also shared with third parties.[12] Healthcare providers, who are often the data controllers and therefore have responsibility for the data they hold, can come into agreements with artificial intelligence (AI) companies to share data for the purposes of developing AI models. Firstly, this can be legally complex and requires contractual controls in place to control the use of data – but it can be achieved alongside the correct ethical and governance framework. Secondly, at the very least, there would be an expectation for financial compensation - in the UK, the second principle of the "Value Sharing Framework for National Health Service (NHS) data partnerships" states that "the NHS should seek to recover the costs of providing access to data. Failing to recover these costs, including a proportionate share of overheads, takes money away from frontline services".[13] Patients and members of the public in the UK would also likely want the NHS to share in the value created by its data.[14] Individual radiologists or researchers are employed by the data controllers and are not the data controllers themselves, therefore have no right to unilaterally share data with commercial organisations.

The reasons mentioned above may explain why the authors found almost no published studies using these three models in similar radiology research. We would therefore encourage any reader of this study to not share even anonymised patient data to ChatGPT or any online service without ethical approval and explicit permission of the data controller (often the hospital or healthcare provider). A practical alternative involves utilising open-source large language models (LLMs) like Llama or its derivatives, which can be set up and run directly on a local computer within a hospital or research organisation's secure network.[15] Here, the research community is active.

From a technical perspective, it seems surprising that the prompt yielded a single-word answer ('yes' or 'no') from all three models, with no other words in the response. In our own experience of LLMs (locally set up on a secure network), extracting the label from a paragraph of the model's reasoning or explanation is an extra cleaning step. We would encourage the authors to share if this was done by human interpreters, or automatically with a script. The poor performances of all models could be explained by the quality of the prompt, which itself should be considered a hyperparameter that can be optimised. A large amount of the LLM literature is dedicated to optimal prompt engineering.[16] Providing the LLM with positive and negative examples, allowing the model to perform 'few-shot' rather than 'zero-shot' inference is an example, and techniques such as chain-of-thought prompting and self-consistency are straightforward methods of increasing performance. It would have also been interesting to have seen a comparison with the auto-encoding type transformer architectures, such as Bidirectional Encoder Representations from Transformers (BERT) and its derivative models.[17,18] These models are naturally suited for classification tasks and can be fine-tuned easily on a subset of the labels to achieve over 95% balanced accuracy for binary abnormality detection in radiology reports.[19,20]